# The Effect of Security-Focused Prompts on Reported Findings: A Controlled Experiment on 400 AI-Generated Web Applications

**Darko Andročec**

University of Zagreb, Faculty of Organization and Informatics, Pavlinska 2, 42000 Varaždin, Croatia; dandrocec@foi.unizg.hr; ORCID: 0009-0007-0180-6566

Abstract: Agentic coding tools turn a specification into a running web application, so their defaults for authentication, sessions, input validation, and dependencies become its security posture. This study asks whether appending a fixed block of OWASP-aligned security requirements to the prompt changes what such tools produce, by how much, and with what side effects. Each of 100 frozen specifications was realized four times: two agentic tools (Claude Code, Codex CLI) crossed with two prompts sharing a byte-identical functional core, without (A) and with (B) the security block. A fixed static and dynamic pipeline scanned the 400 applications, yielding 6,721 triaged findings, analyzed with count and logistic mixed models and with a dependency diff. The block cut findings per application by a third (incidence-rate ratio 0.65, 95% CI 0.60–0.71), equally through both tools, and roughly doubled the declared dependency surface, so variant-B applications carried more high-severity dependency advisories. Adjusted for that surface the advisory rate did not rise (flat per application, 18% lower in aggregate), while high-severity findings in first-party source fell by nearly half. Security-focused prompting is therefore an inexpensive and portable improvement, but it enlarges the dependency surface that has to be audited against it, an effect of library selection rather than of the security requirements themselves.

# 1 Introduction

Agentic coding tools are command-line agents built on large language models that plan, write files, install dependencies, and iterate until an application runs. They have shifted the unit of AI code generation from the autocompleted line to the complete, deployable system, and with it the security question: when an agent delivers an entire web application, the defaults it chooses for authentication, session handling, input validation, and dependency selection become the application's security posture, frequently without any human reading the relevant lines. The practice is spreading under the label of “vibe coding”, an iterative cycle of prompting, rapid appraisal of the output, and selective manual repair [18], and applications built this way already show characteristic weaknesses that stem from limitations of the generating agent rather than from any individual coding error [5]. Web applications are a particularly consequential setting: they are exposed to the public internet, they handle authentication, personal data, and payments, and they sit atop deep dependency trees in which a single manifest line can pull dozens of transitive packages, and any published vulnerability among them, into the deployed artifact [13, 26].

Decades of effort captured in the OWASP Top 10 [14] show that most real-world web compromises exploit well-understood, preventable weakness classes, and one countermeasure is available at negligible cost: stating security requirements in the prompt itself. That such an instruction reduces the number of weaknesses a scanner reports is not, in itself, surprising, and it is not what motivates this study. Three properties of the intervention are unknown and consequential: its magnitude, since the advice is worth issuing only if the effect is large; its uniformity, since an effect that depends on which agent a team has adopted cannot become a portable practice; and its side effects, since a security requirement is not a filter applied to a finished application but an input to a generative process that decides what to build and what to install, so it may change the artifact twice over, removing weaknesses while enlarging it in ways that carry risk of their own. Snippet-level studies cannot observe this, because the behaviors at stake exist only at the whole-application level, among them choosing a password-hashing library, wiring CSRF middleware, and pinning dependency versions.

The objective is therefore to quantify the effect of a security-focused prompt on the findings reported for complete, runnable web applications, and to determine what else that instruction changes in the delivered artifact. One hundred frozen specifications were each realized four times, by two independent agentic tools, each prompted with and without a fixed OWASP-aligned security-requirements block, and the 400 resulting applications were scanned by an identical pipeline. The functional portion of every prompt was byte-identical across the four conditions of a specification, so any systematic difference in scan results is attributable to the tool, the security block, or their interaction. Six research questions structure the analysis. RQ1: does the security block reduce the number of findings per application? RQ2: do the two tools differ in the number of findings they produce? RQ3: how do findings scale with task complexity and language ecosystem? RQ4: does the block change the severity composition rather than only the volume? RQ5: does the block's effect depend on the tool? RQ6: does it depend on task complexity?

The study makes three contributions. First, it characterizes the instruction's effect on the delivered artifact: it approximately doubles the declared dependency surface, frequently adding authentication and session machinery to applications whose functional core has no users, and the accompanying rise in high-severity dependency advisories is accounted for by that enlarged surface rather than by any deterioration in the generated code. Second, it localizes that exposure in library selection rather than in the security requirements themselves. Third, it is, to the author's knowledge, the largest controlled comparison of security-focused against neutral prompting over complete applications generated end-to-end by agentic tools, and the first to cross that instruction with two independently developed tools.

A preliminary version of this work has been accepted as a short conference paper, to be presented in September 2026 and available as a preprint [2]; it introduces the twin-prompt idea and applies it to twelve manually adjudicated programs from a single assistant, reporting descriptive observations only. The present study is the announced scaling of that pipeline: the corpus is two orders of magnitude larger and frozen under a SHA-256 lock file before generation, the design is crossed with two independently developed tools, the analysis is inferential, and the dependency analyses of Sections 4.4 and 4.5 have no counterpart in it. One of its conclusions is revised here: in twelve programs the security-aware variant contained no high-severity issue, whereas across 400 applications the same instruction is associated with *more* high-severity findings in absolute terms, because it enlarges the dependency surface that software-composition analysis examines. No text, table, figure, or result from that paper is reused here.

# 2 Related Work

## 2.1 Security of AI-Generated Code

The first strand of related work audits the code that generators emit. Pearce et al. [16] prompted GitHub Copilot with 89 scenarios built around MITRE's most dangerous CWEs and found approximately 40% of the 1,689 generated programs vulnerable. Dora et al. [6] narrowed the target to the setting of this study, reporting for web application code from five widely used assistants deficiencies concentrated in authentication, session management, input validation, and HTTP security headers. In the field, Fu et al. [7] detected weaknesses in 29.5% of Python and 24.2% of JavaScript Copilot-generated snippets in public GitHub projects. Closest in setting, Deng et al. [5] examined applications produced by agent-driven development and, in 360 controlled runs varying agent harness, model, and user prompt, found lightweight prompt-engineering variants the most effective mitigation. A second strand studies the developer rather than the artifact: Perry et al. [17] found that participants solving security-relevant tasks with an AI assistant wrote less secure code while being more likely to believe it secure, which establishes the premise of the present work, that if the human in the loop does not supply the security requirement, the generator's default is what ships. A third strand builds evaluation corpora, two of which operate at the level of whole programs: BaxBench, where roughly half of the functionally correct generated backends remained exploitable [22], and SecRepoBench, with 318 secure-completion tasks inside real C/C++ repositories [20].

## 2.2 Steering Generation Towards Security

The fourth strand intervenes on generation itself. He and Vechev [8] learned continuous prompt prefixes that harden, or adversarially weaken, a code model's output without modifying its weights, an intervention that requires access to the model and is therefore unavailable to a practitioner who prompts a hosted agent, which is the situation this study measures. Tony et al. [21] report that security-focused prompting refinements substantially reduce reported weaknesses in Python generated by GPT models, and He et al. [9] pose the baseline-against-security-oriented-prompt comparison in terms closest to this paper's own, over ten single-file Python tasks on two models, finding that security-informed prompts lowered both the frequency and the severity of reported weaknesses but not uniformly across models. The evidence is mixed: Kharma et al. [11] found across five models and four languages that prompting shifts the distribution of CWE categories produced but yielded no statistically significant reduction in overall vulnerability counts, and in BaxBench a generic security reminder helped reasoning models but not others [22]. Whether a practitioner-style security instruction helps, and how much, is therefore an open empirical question.

## 2.3 Dependencies and Measurement Instruments

The applications studied here are ordinary npm, PyPI, and Composer applications, and the severity results of Section 4.4 are governed by established ecosystem properties: an npm package implicitly trusts, on average, 79 third-party packages and 39 maintainers [26], counting vulnerability propagation correctly requires resolving dependency trees by the registry's actual resolution rules [13], and install scripts are a known weak-link signal across 1.63 million npm packages [24] and are the mechanism by which the native build chains of Section 4.5 enter an application. One further result bears on how the severity analysis must be read: advisory counts are a function of what a project declares, so counting them without regard to actual use over-states exposure [15]. To the author's knowledge, no prior study has measured how a security instruction in a prompt changes that surface.

The outcome measure is what a fixed pipeline of analyzers reports, and their properties condition every estimate: false positives are the primary barrier to the adoption of static application security testing [10], rule-based and learning-based approaches show low agreement and modest precision on real code [4], out-of-the-box rule sets determine what is reported in practice because most developers do not tune such tools [3], and web vulnerability scanners show precision and recall well short of complete [1]. A findings count is therefore a noisy proxy whose absolute level is not interpretable, but the same pipeline, rule sets, and version pins are applied to all 400 applications, so that bias is held constant across the manipulated factor.

### 2.4 Positioning of This Study

A targeted search of the ACM Digital Library, IEEE Xplore, Springer Link, and arXiv, conducted in August 2026 with backward and forward snowballing, identified two studies that manipulate a security-relevant instruction as a controlled factor over complete, runnable applications generated end-to-end by agentic command-line tools: the author's preliminary twelve-program study [2], and the fourth research question of Deng et al. [5], which varies prompt framing alongside the model and the harness for one agentic tool and scores whether a known vulnerability reappears in a reproduction task. The prompting studies [8, 11, 21] operate on snippets or single functions with the vulnerable scenario fixed by the researcher, and the application-level benchmarks [20, 22] generate whole backends or repository-level completions by single-shot prompting rather than by an agent that plans, writes multiple files, installs dependencies, and iterates. SusVibes [25] is agentic throughout and does vary a security-relevant instruction, but its 186 tasks are feature requests against existing projects and the instruction it varies is a vulnerability hint rather than a requirements block. BaxBench is larger in generated programs, evaluating 392 tasks across 11 models with and without a generic security reminder [22], but it issues a single-shot prompt for one predominantly single-file backend per task in a prescribed framework. Three further properties distinguish the present design: the specification, not the model, is the experimental unit and is realized in all four conditions, so the instruction is compared against a byte-identical functional core; two independently developed tools are crossed with the instruction, which makes portability estimable rather than assumed; and the outcome is measured over the delivered artifact as a whole, including the dependency manifest the agent writes.

## 3 Method

### 3.1 Design and Specification Catalog

The experiment follows the process and terminology of Wohlin et al. [23]. It used a fully crossed 2 × 2 × 100 design: two agentic generation tools (claude, codex) × two prompt variants (A, B) × one hundred application specifications, yielding 400 generated applications. The specification is the experimental unit and tool and variant are within-specification factors; the specification enters the analysis as a random intercept in the mixed models and as the clustering variable for robust standard errors in the fixed-effects models. The catalog size was fixed before generation on grounds of yield rather than a power calculation, since it produces 200 within-specification A/B pairs, which is the quantity the design is built to exploit; no effect size was posited in advance and no formal power analysis was performed. The catalog was authored and frozen, with a SHA-256 lock file, before any generation began. Specifications span four complexity tiers, from single-purpose utilities (T1, $n$ = 25) through small CRUD services (T2, $n$ = 30) and multi-entity applications (T3, $n$ = 30) to platform-style systems with roles and workflows (T4, $n$ = 15), and three language ecosystems: Node.js with Express ($n$ = 46), Python with Flask or FastAPI ($n$ = 39), and plain PHP ($n$ = 15). Each specification fixes the language, framework, TCP port, and a list of functional requirements.

## 3.2 Prompt Variants and Generation

Two prompt variants were rendered mechanically from each specification. Variant A consisted of the specification's functional core plus a neutral footer asking for a runnable application with a dependency manifest and a README. Variant B was identical, and byte-identical in its functional core, except for one inserted paragraph of mandatory security requirements, itself byte-identical across all 200 variant-B prompts. Figure 1 reproduces both prompts verbatim for one specification.

*Variant A (specification S001, rendered prompt, verbatim)*

Build a tiny web app with Python and Flask. Show a form with a single text field for a person's name and a submit button. When submitted, display a page that greets the user by the name they entered (for example, "Hello, <name>!"). No database is needed.

Make the application fully runnable; provide the appropriate dependency manifest (requirements.txt / package.json / composer.json) and a short README explaining how to run it locally on port 5001.

*Variant B = variant A, unchanged, plus exactly this appended paragraph*

Security requirements (mandatory). Apply OWASP Top 10 best practices throughout. Use parameterised queries to prevent SQL injection. Hash passwords with a strong, salted algorithm (bcrypt or Argon2). Validate and sanitise all user input and apply context-aware output encoding to prevent XSS. Implement CSRF protection on all state-changing requests. Enforce proper access control so users can act only on their own resources (prevent IDOR). Configure secure session cookies (HttpOnly, Secure, SameSite). Set appropriate security headers. Do not leak stack traces or internal errors to clients. Do not hardcode secrets; read them from environment variables.

Figure 1
The two prompt variants, verbatim; the functional core and the appended security block are each byte-identical across conditions

One property of this design governs how the severity results may be read: the block prescribes concrete mechanisms rather than an outcome, so requirements such as secure session cookies, password hashing, and per-user access control invite authentication machinery even into specifications whose functional core has no users. Variant B therefore manipulates instruction content and implementation scope together; it remains the intervention a practitioner would apply, but it is not a comparison in which the application is held constant, and Section 4.4 quantifies this scope growth and adjusts for it.

Applications were generated in clean, isolated sessions by Claude Code (Anthropic; model claude-opus-4-8; CLI 2.1.169–2.1.186) and Codex CLI (OpenAI; model gpt-5.5; version 0.140.0), each invoked non-interactively with equivalent commands and recorded in an append-only command log. Generated code was treated as immutable: no bug fix, hardening, or reformatting was applied after generation, and build failures were addressed only by re-running generation into a clean directory. The Claude Code CLI advanced through ten releases over the generation window whereas Codex CLI stayed at 0.140.0, but the two variants of a given specification were generated back to back, so 196 of the 200 specification × tool pairs were produced by an identical CLI version and the variant contrast is estimated within CLI version almost everywhere. Generation took 27.8 hours of agent time, a median of 175 seconds per application, the security block raising that median from 180 to 350 seconds for claude and from 129 to 168 seconds for codex. Each application was then built and started in Docker to verify that it runs; 386 of the 400 passed, and the 14 genuine generation defects were retained unmodified as data points.

## 3.3 Analysis Pipeline

All 400 applications were analyzed by the same pipeline, with every scanner in its default configuration, which is what a majority of practitioners do [3], and no rule enabled or suppressed for any individual application. Static analysis used *bandit* and *pip-audit* (Python), ESLint with the security plugin and *npm audit* (Node.js), *progpilot* (PHP), plus *semgrep* across all languages. Dynamic analysis comprised a passive OWASP ZAP baseline scan and an active pass with Nikto and targeted *sqlmap* probes, run against each application executing in a deliberately security-neutral container harness, with no added headers, hardening, capability drops, or non-root user, so that every observation is attributable to the generated code. Static analysis covered all 400 applications; dynamic analysis covered the 392 reachable over HTTP, eight applications having failed to boot, five under variant A and three under variant B, so the resulting loss of dynamic findings runs against, rather than produces, the lower counts observed under variant B.

Two properties of this pipeline are asymmetric across the three language ecosystems and must be stated before any language comparison is read. First, *npm audit* and *pip-audit* query advisory databases for the packages a manifest declares whereas *progpilot* is a static taint analyzer, so no software-composition analysis was performed for PHP and PHP applications cannot by construction yield a finding in the Vulnerable and Outdated Components category (A06), which accounts for 1,471 of the 6,721 findings and 542 of the 699 HIGH findings. Second, *progpilot* contributed 4 findings in total across the 60 PHP applications, leaving *semgrep* as effectively the only static instrument there, so static findings per application are 7.3 (Node.js), 7.0 (Python) and 0.6 (PHP), whereas dynamic analysis is equivalent for all three ecosystems.

## 3.4 Dataset Construction and Statistical Analysis

Scanner reports were transcribed into a single per-finding dataset, copying only what each tool asserted. Findings reported by multiple scanners within one run were merged, and a conservative triage flagged, without deleting, finding classes known to be artifacts of this harness: catch-all routing echoed by Nikto path probes, ZAP informational fingerprints, and bind-to-all-interfaces inside a container. Of 7,553 consolidated findings, 832 were flagged as likely false positives, leaving an analysis population of 6,721. These are tool-reported findings, not adjudicated vulnerabilities; given the known imprecision of security analyzers [4, 10], no finding was confirmed exploitable, and effect estimates should be read as associations over reported findings.

Severity was copied from the reporting scanner and never assigned by the author, which leaves the population distributed over five labels rather than the conventional four: LOW (3,171), MEDIUM (1,214), INFO (961), HIGH (699) and DEPENDENCY (676). No scanner asserted a CVSS score, so severity is an ordinal label. INFO comprises the informational alerts of the passive ZAP scan. DEPENDENCY is not a severity band but an artifact of scanner output, comprising exactly the 676 *pip-audit* advisories, all of them Python and all of them A06, because *pip-audit* reported known vulnerable packages without a severity field; rather than impute one, the label records its absence. That asymmetry matters, because *npm audit* does assert severities, its 795 advisories dividing into 542 HIGH, 248 LOW and 5 MEDIUM, so a dependency advisory can be classified high severity in the Node.js ecosystem and cannot be so classified in the Python one for reasons of instrument output alone. High severity is operationalized as HIGH or CRITICAL, and no CRITICAL occurred; DEPENDENCY and INFO findings consequently enter the logistic model in the reference category by construction, and Section 4.3 reports a sensitivity analysis that removes them.

To interpret the severity results (RQ4), a dependency-diff analysis was added. The direct dependencies declared in every manifest were extracted, and within each specification × tool pair the variant-B set was compared against variant A. Because software-composition analysis reports advisories in proportion to what a manifest declares [15], each application's declared dependency surface was recorded so that high-severity advisory counts could be expressed per declared direct dependency and not only per application. Scope growth beyond the functional core was measured separately, by identifying from the frozen catalog the specifications whose functional core mentions no users, accounts, roles, or sessions and checking their applications for declared authentication or session packages. Every high-severity *npm audit* advisory was then traced through its dependency chains to the manifest package that pulls the vulnerable package into the build, and classified by whether that anchor occurs only in the variant-B manifest or is shared with the paired variant-A application. This analysis is read-only over the frozen artifacts and raw scanner reports.

The per-application finding count ($n$ = 400 rows) was modeled with a negative-binomial generalized linear model using cluster-robust standard errors by specification, and a Poisson generalized linear mixed model (random intercept for specification, variational-Bayes fit) was fitted in parallel as a dispersion check. Fixed effects were prompt variant, tool, tier, language, and the variant × tool interaction, with reference levels A, claude, T1, and Node.js; for RQ6 the count model was refitted with a variant × tier interaction added, the three interaction coefficients being tested jointly rather than read off individually. Severity composition was modeled at the finding level with a logistic mixed model, and pairwise contrasts were formed as linear combinations of the fixed-effects coefficients. Incidence-rate ratios (IRRs) and odds ratios (ORs) are reported with 95% intervals. Two models were fitted for each outcome because agreement between estimators with different assumptions is a more defensible basis for a conclusion than any single $p$-value [12]. Analyses used Python 3.14 and statsmodels 0.14.6 [19].

# 4 Results

The analysis population contained 6,721 findings over 400 applications: a mean of 16.80 findings per application (SD 6.12, range 3–44). Volume was driven almost entirely by the prompt variant, not the tool: mean findings per application were 20.15 (SD 5.91) for claude and 20.04 (4.56) for codex under variant A, falling to 13.18 (4.87) and 13.84 (5.23) under variant B, while the HIGH counts per cell ran 154 and 147 under variant A against 168 and 230 under variant B. By detection method, the passive dynamic scan contributed 3,463 findings (51.5%), static analysis 2,478 (36.9%), and the active dynamic scan 780 (11.6%). Of the 699 HIGH findings (10.4% of the population) every one came from static analysis and none was CRITICAL. Two OWASP Top 10 categories account for the majority of findings, Security Misconfiguration (A05, 2,503) and Vulnerable and Outdated Components (A06, 1,471).

## 4.1 Finding Volume and Tool Comparison (RQ1, RQ2, RQ5)

Adjusting for tool, tier, and language, variant B was associated with 35% fewer findings per application than variant A (negative-binomial IRR 0.65, 95% CI 0.60–0.71), and the Poisson mixed model agreed almost exactly (IRR 0.65, 95% CI 0.63–0.68), indicating that the effect is not an artifact of between-specification heterogeneity. The tools did not differ in overall finding volume (codex vs. claude IRR 0.99, 95% CI 0.94–1.06), and the variant × tool interaction was small and non-significant (IRR 1.06, 95% CI 0.96–1.16), so the security block was approximately equally effective through both tools, reducing counts by 35% within claude (IRR 0.65, 95% CI 0.60–0.71) and by 31% within codex (IRR 0.69, 95% CI 0.64–0.74). Within each variant separately the tools remained statistically indistinguishable, at IRR 0.99 (95% CI 0.94–1.06) under variant A and 1.05 (0.98–1.13) under variant B. Table 1 reports the full model.

Table 1
Model estimates. IRR: negative-binomial model of findings per application ($n$ = 400, cluster-robust by specification). OR: logistic mixed model of high severity ($n$ = 6,721, random intercept for specification). Reference levels A, claude, T1, Node.js; language rows are covariates only (Section 4.2)

| **Term** | **IRR (95% CI)** | **OR (95% CI)** |
|---|---|---|
| Intercept | 18.55 (17.20–20.00) | 0.07 (0.06–0.07) |
| Variant B (vs. A) | 0.65 (0.60–0.71) | 1.71 (1.53–1.91) |
| Tool codex (vs. claude) | 0.99 (0.94–1.06) | 0.92 (0.82–1.03) |
| Tier T2 (vs. T1) | 1.21 (1.12–1.31) | 3.09 (2.70–3.54) |
| Tier T3 (vs. T1) | 1.15 (1.06–1.24) | 2.71 (2.36–3.09) |
| Tier T4 (vs. T1) | 1.32 (1.18–1.49) | 2.04 (1.68–2.48) |
| Language PHP (vs. Node.js) | 0.66 (0.63–0.71) | 0.07 (0.04–0.13) |
| Language Python (vs. Node.js) | 0.97 (0.91–1.03) | 0.21 (0.18–0.26) |
| Variant B × tool codex (interaction) | 1.06 (0.96–1.16) | 1.40 (1.21–1.62) |

## 4.2 Complexity and Language Gradients (RQ3, RQ6)

Findings rose with task complexity: relative to the simplest tier, IRRs were 1.21 (T2), 1.15 (T3) and 1.32 (T4), and mean findings per application climbed from 14.1 (T1) to 20.2 (T4). Whether the effect of the security block is constant across those tiers (RQ6) cannot be settled by the model of Table 1, which carries no variant × tier term, so it was refitted with one added. No single interaction coefficient reaches significance (B × T2 IRR 0.93, 95% CI 0.83–1.06; B × T3 IRR 1.12, 95% CI 0.96–1.30; B × T4 IRR 1.18, 95% CI 0.96–1.44), and although the three are jointly distinguishable from zero, a joint test resting on individually inconclusive terms and not adjusted for the several contrasts examined here is weak evidence rather than a demonstration. Read cautiously, the data indicate a weakening of the effect in the more complex half of the catalog without establishing it: the block reduced findings significantly in every tier (all $p < 0.01$), but the estimated reduction was larger in the two simpler tiers, at IRRs of 0.63 (T1) and 0.59 (T2) against 0.70 (T3) and 0.74 (T4). The tier-specific intervals overlap and the pattern is not monotone, so it is better read as a difference between the two halves of the catalog than as a gradient. It suggests that one fixed paragraph of requirements buys proportionally less as the application grows, which is consistent with what the block is observed to remove, namely application-level configuration and hygiene, a component that does not scale with the size of the functional core; security outcomes are likewise worse in repository-scale settings than in self-contained ones [20].

The models also return language coefficients, but these are not results about the languages. PHP applications yielded about a third fewer findings than Node.js applications (IRR 0.66, 95% CI 0.63–0.71), Python not differing from Node.js, and a PHP finding had far lower odds of being high severity (OR 0.07, 95% CI 0.04–0.13). The deficit falls where the coverage asymmetry of Section 3.3 lies, in static analysis, whereas dynamic findings, produced by identical instruments, are near-identical at 10.7 (Node.js), 10.9 (Python) and 10.8 (PHP) per reachable application. The language terms are therefore covariates only, and no comparison between languages is claimed.

## 4.3 Severity Composition in Unadjusted Terms (RQ4)

Volume and unadjusted severity composition moved in opposite directions. Although variant B produced far fewer findings overall (A: 4,019; B: 2,702), it produced more HIGH findings in absolute terms (A: 301; B: 398), and in the logistic mixed model of Table 1 a finding generated under variant B had higher odds of being high severity (OR 1.71, 95% CI 1.53–1.91); the fixed-effects logistic fit with cluster-robust standard errors gives an almost identical point estimate on a considerably wider interval (OR 1.72, 95% CI 1.23–2.40), and where the two error models disagree the wider interval is relied on here. Read consistently from that fit, the shift holds within both tools (claude OR 1.72, 95% CI 1.23–2.40; codex OR 2.38, 95% CI 1.85–3.07), while the codex-against-claude contrast within variant B, at OR 1.28 (1.00–1.64), should not be read as established. Because 1,637 findings carry the INFO or DEPENDENCY labels and so count as not high severity by construction, the model was refitted on successively narrower populations, which moves the estimate away from unity rather than towards it: 1.77 (1.58–1.98) with DEPENDENCY excluded and 2.02 (1.80–2.26) with DEPENDENCY and INFO both excluded. The estimate reported above is therefore the most conservative of the three.

Splitting the HIGH findings by origin locates the shift precisely. High-severity findings in the applications' own source code (*bandit*, *semgrep*) fell from 103 under variant A to 54 under variant B, while high-severity dependency advisories rose from 198 to 344. At the level of first-party code the security block improved the high-severity outcome; the increase is confined to the dependency layer, and of the 699 HIGH findings 542 fall under A06. This model does not adjust for how much dependency surface each application exposes to software-composition analysis; Section 4.4 supplies that adjustment, and it removes the difference.

## 4.4 Dependency Surface Growth Accounts for the Severity Shift

The dependency-diff analysis shows that variant B changes the dependency surface itself, not merely the configuration of an otherwise identical application. Complying with the security block approximately doubled the declared footprint: mean direct dependencies per application rose from 3.2 to 6.3 (claude) and from 3.0 to 6.5 (codex), with 171 of the 200 specification × tool pairs gaining at least one new direct dependency. Within-pair comparisons rest on the 86 specification × tool pairs whose Node.js applications produced a manifest under both variants, a population holding 526 of the 542 high-severity npm advisories; there all 86 pairs declared more direct dependencies under variant B, the mean rising from 4.49 to 9.87. Part of that growth lies outside the functional core and is therefore declared, installed, audited, and unnecessary: of the 46 Node.js specifications, 20 describe an application with no users, accounts, roles, or sessions of any kind, and across the 38 comparable pairs built from those specifications 36 (95%) declared an authentication or session package under variant B against 17 (45%) under variant A. Requirements phrased as mechanisms are evidently read by the agents as features to implement, not only as constraints on features already requested.

The surplus advisories concentrate accordingly in a few build-heavy packages that variant B introduces. Of the 344 high-severity npm advisories in the 87 manifest-bearing variant-B applications, 124 also occur in the paired variant-A application and the remaining 220 trace exclusively to packages absent from the variant-A manifest. Three anchor packages carry all of them, whose native build chains (*tar*, *node-gyp*, *node-pre-gyp*, *make-fetch-happen*, *cacache*) carry the published advisories: the session store *connect-sqlite3* (added in 34 specification × tool pairs, 170 advisories), the password-hashing library *bcrypt* (58 pairs, 116 advisories), and the database driver *sqlite3* (8 pairs, 40 advisories). This is the propagation pattern the npm literature describes, in which the vulnerable package is typically not the one the developer named but one reached transitively through it [13], and packages with install scripts and native build steps are precisely the class previously identified as a weak-link signal in that ecosystem [24]. The reverse flow is smaller: 74 advisories occur only under variant A.

Once this exposure is taken into account, the difference between the variants disappears. All figures in this paragraph and in Table 2 are computed on the paired population of 86 applications per variant. High-severity npm advisories per application rose 1.79-fold under variant B while the declared dependency surface rose 2.20-fold, so the advisory rate per declared direct dependency did not increase: in aggregate it was 0.487 under variant A against 0.398 under variant B, and 0.366 against 0.371 when computed per application and then averaged. The additional advisories are what a larger declared surface mechanically yields when the same software-composition analysis is applied to it, the counting effect that Pashchenko et al. [15] identified as the principal source of over-statement in dependency-vulnerability reporting. They are not evidence that security-focused prompting produced more severely vulnerable code, since within first-party source the same instruction approximately halved high-severity findings. The severity results of Section 4.3 should therefore be read as a statement about how much dependency surface each variant exposes to a scanner, not as a reversal of the effect measured in Section 4.1.

Table 2
Exposure-adjusted high-severity npm advisories, over the 86 specification × tool pairs with a manifest under both variants

| **Quantity** | **Variant A** | **Variant B** | **Ratio B/A** |
|---|---|---|---|
| Declared direct dependencies per application (mean) | 4.49 | 9.87 | 2.20 |
| HIGH npm advisories per application (mean) | 2.19 | 3.93 | 1.79 |
| HIGH advisories per declared dependency (aggregate) | 0.487 | 0.398 | 0.82 |
| HIGH advisories per declared dependency, per-application mean | 0.366 | 0.371 | 1.01 |

## 4.5 Evidence on Dependency-Specific Instructions

The security block names capabilities but no packages. Some agents chose packages whose current versions build native extensions, namely *bcrypt*, *connect-sqlite3* and *sqlite3*; others chose equivalents that satisfy the identical requirement without a build toolchain, namely *bcryptjs*, *argon2* and *better-sqlite3*. The distinction is exactly the install-script signal that Zahan et al. [24] identified as a weak link in the npm supply chain, arrived at here by agent choice rather than by developer policy. The two package lists were assembled after the dependency-diff analysis had identified the three anchor packages, so what follows is a post-hoc partition of an observed corpus and is hypothesis-generating; it is reported because it bears on a question the experiment cannot otherwise answer, namely whether extending the prompt with dependency-specific instructions would help.

The separation is complete. Of the 79 manifest-bearing variant-B Node.js applications that declared at least one of these capabilities, the 12 that declared only chain-free equivalents carried no high-severity advisories whatever, while all 67 that declared at least one chain-heavy package carried some, a mean of 5.1 each. Neither application size nor tool identity accounts for the contrast: the chain-free applications declare slightly fewer direct dependencies rather than more, a mean of 9.1 against 10.6; nor are they the simpler ones, since 5 of the 12 come from the most complex tier T4 against 10 of the 67; and within claude alone, which produced 11 of the 12, the difference is 0.00 against 5.17 mean advisories, with no overlap between the groups. The reading this supports is that the high-severity exposure introduced under variant B is attributable to library selection rather than to the security requirements themselves, and that naming chain-free equivalents in the prompt would remove most or all of it while leaving every requirement of the block satisfied. The estimate is bounded by its observational nature, since library choice was not randomized, it correlates with tool, and twelve applications is a small base.

# 5 Discussion

A single fixed paragraph of security requirements, adding only prompt tokens to the cost of generation, was associated with approximately a third fewer security findings per generated application. The effect replicated across two independently developed tools and was significant in all four complexity tiers, though its estimated magnitude was not constant: roughly a 40% reduction in the two simpler tiers against roughly 26 to 30% in the two more complex ones, an attenuation with which the joint interaction test is consistent but which it does not establish. Most of the reduction occurred in the classes a passive scanner detects, namely missing security headers, cookie attributes, and verbose defaults, and that composition offers a plausible mechanism for the apparent attenuation, since these are largely fixed-cost properties of an application, so the share of findings the block can remove falls as the functional core grows. When instructed, both tools close a hygiene gap that they otherwise leave open, which raises the question of why such behavior is not their default.

The result sits between two positions in the prompting literature. Studies reporting substantial gains from security-focused prompting [21] and studies reporting no significant reduction in vulnerability counts [11] both operate on snippets, and BaxBench found a generic security reminder to help reasoning models but not others [22]. The present estimate is positive, moderate, and uniform across two tools, but it is measured over a different class of weakness, since what improves is predominantly application-level configuration and hardening, which has no counterpart in a single generated function. A security instruction thus appears to be effective mainly at the level at which whole-application decisions are made, and correspondingly weak where prior snippet-level work has looked for it. SusVibes marks the boundary of the claim: on 186 feature-request tasks in real open-source projects only 11.8% of the strongest agent's solutions are secure, and vulnerability hints do not mitigate the problem [25]. Their agents change an existing repository, where the configuration and hygiene layer is already fixed, whereas the applications measured here are generated from nothing, so the instruction arrives at the moment the defaults are chosen. A security-requirements block should therefore be expected to help where it can set defaults, and not to substitute for review where an agent edits logic inside an existing codebase.

The severity results require more care, because variant B does not hold the application constant: the block prescribes mechanisms, the agents implement them as features, and software-composition analysis then reports advisories in proportion to what the enlarged manifest declares. The block thus relocates risk rather than creating it, shrinking the configuration attack surface while enlarging the dependency attack surface. The advisories reside not in the added libraries' own code but in their native build chains, so the generated source contains no visible defect and only manifest-level analysis reveals the exposure, and an enlarged dependency tree also widens the set of packages and maintainers a project implicitly trusts [26]. An agent that silently adds five packages to satisfy a security requirement is making a trust decision on the developer's behalf, in the part of the artifact that developers are least likely to read. Three low-cost correctives follow: security requirements should be scoped to the application's trust model rather than applied wholesale; the prompt should be paired with dependency auditing in the build pipeline; and the libraries should be named, not only the capabilities. The last is the most promising and the least explored, since the block leaves the choice of package to the agent and that choice is where the high-severity exposure resides.

## 5.1 Threats to Validity

*Construct validity.* Outcomes are tool-reported findings, not adjudicated exploitable vulnerabilities, so the measure captures what a standard pipeline reports rather than realized risk, and the instruments are known to be imprecise in both directions [1, 4, 10]. Two design choices bound the consequence: multiple independent scanners are combined rather than relied upon singly, and the entire pipeline, including default rule sets [3] and version pins, is identical across all 400 applications, so instrument bias is held constant across the manipulated factor and differences between conditions remain interpretable even though absolute levels are not. What cannot be excluded is an interaction between instrument and treatment. The pipeline's coverage is also not equivalent across the three ecosystems, so the language coefficients measure instrument coverage rather than any property of the generated code; and counts pool three detection methods without adjustment, while dependency advisory counts partly measure declared surface rather than code quality [15], which is why the exposure-adjusted analysis is reported alongside the raw counts.

*Internal validity.* The most consequential threat is that variant B confounds security semantics with implementation scope: its prompt is longer and more prescriptive, and it demonstrably causes more functionality and more dependencies to be produced, so the design cannot attribute either the reduction in findings or the change in severity composition to security wording alone. A placebo instruction block of matched length, and a variant requiring the same protections while forbidding new direct dependencies, would separate instruction volume, security semantics, and implementation scope. The library-choice comparison of Section 4.5 is observational rather than randomized and correlates with tool, and each cell contains a single generation per specification, tool, and variant, so run-to-run stochasticity is absorbed into the residual rather than estimated.

*External validity.* The results are conditional on two agentic tools, one hundred specifications, three language ecosystems, and the specific scanners studied. The applications are small-to-medium web applications rather than large systems, and the specifications were authored by the researcher rather than sampled from industry backlogs.

*Conclusion validity.* The random-intercept count model relied on a variational-Bayes Poisson approximation whose intervals may be anti-conservative, although the fixed-effects model with cluster-robust standard errors produced the same estimates, and agreement between two estimators with different assumptions is the basis on which the effect is reported [12]. Pairwise contrasts were not adjusted for multiplicity, so the individual intervals should be read as descriptive; the same caution applies with more force to the variant × tier interaction (RQ6), which is reported because the pattern is orderly and practically consequential rather than because it is established.

## Conclusions

Agentic coding tools are becoming a routine part of software development practice, and a growing share of deployed web applications will contain code that no developer authored directly, so whatever systematic weaknesses the generators exhibit will be replicated at the scale and speed at which the tools operate. This study examined one concrete and low-cost measure available to every practitioner, the wording of the prompt itself. In a controlled experiment over 400 AI-generated web applications, an explicit OWASP-aligned security instruction reduced reported security findings by about one third, uniformly across two agentic coding tools that were otherwise statistically indistinguishable. The same instruction approximately doubled the declared dependency surface, so applications built under it carried more high-severity dependency advisories in absolute terms; adjusted for that surface the advisory rate per declared dependency was unchanged, and high-severity findings in first-party source fell by nearly half.

The results carry a practical message with two parts. Security-aware prompting is among the least costly interventions available in the AI-assisted development workflow, and a single fixed paragraph of requirements measurably improved the hygiene of the generated applications across tools, complexity tiers, and language ecosystems, which argues for embedding such instructions into organizational prompt templates, agent configuration files, and coding guidelines as a matter of default practice. Requesting security features also delegates dependency selection to the tool, and the exposure that follows is proportional to what is installed rather than a reversal of the security benefit, but it is genuine and it is not visible in the generated source. Four practices therefore follow: security should be requested explicitly; the request should be scoped to the application's requirements; the libraries to be used should be named rather than only the properties they must satisfy; and whatever the tool installs should be audited. Prompt-level security guidance is a complement to, never a substitute for, software-composition analysis and dependency auditing in the build pipeline.

Several directions follow from the design and its limitations. The most valuable is a third generation arm in which the security block carries an explicit dependency-hygiene clause naming chain-free libraries such as *argon2*, *bcryptjs*, or *better-sqlite3*; Section 4.5 predicts that such a clause removes essentially all of the high-severity dependency exposure while leaving the required protections intact, a prediction sharp enough to be refuted by a single further run over the same hundred specifications. Two additional variants would resolve the confound identified above, a placebo block of matched length and a variant mandating the same protections while forbidding new direct dependencies. Adjudicating a sample of findings would estimate the pipeline's true-positive rate and let exploitability serve as the endpoint, and a pipeline with equivalent coverage in every ecosystem would make the language factor estimable. Iterative regimes in which scanner output is fed back to the agent may also achieve reductions unattainable with a single static instruction.

## Acknowledgement

This research was funded by the National Recovery and Resilience Plan 2021–2026, Institutional Research Projects, under the project PiPi: Data infrastructure and augmented intelligence for sustainable development. The author declares no competing interests. The author further discloses that AI tools were the object of study in this work: Claude Code (Anthropic; model claude-opus-4-8) and Codex CLI (OpenAI; model gpt-5.5) generated the 400 web applications analyzed here, under the controlled conditions described in Section 3, and AI-assisted tooling was used during preparation of the manuscript for language editing and the scripted analysis of the frozen artifacts. The author reviewed and verified all AI-assisted outputs against the underlying data, made all analytical and interpretive judgments, and takes full responsibility for the content. AI tools do not meet the criteria for authorship and none is listed as an author.

## Data Availability

The 400 generated applications are preserved unmodified as frozen artifacts, together with the specification catalog and the rendered prompts at the publicly available GitHub repository at the following link: https://github.com/dandrocec/claude_codex_webapps_security/.